\documentclass[a4paper]{article}
\usepackage{INTERSPEECH2022}
\usepackage{adjustbox}
\usepackage{caption}
\usepackage{subcaption}
\usepackage{multirow}
\usepackage[hyphens,spaces]{url}

\title{CoCA-MDD: A Coupled Cross-Attention based Framework for Streaming Mispronunciation Detection and Diagnosis}
\name{Author Name$^1$, Co-author Name$^2$}
\name{Nianzu Zheng, Liqun Deng, Wenyong Huang, Yu Ting Yeung, \\
Baohua Xu, Yuanyuan Guo, Yasheng Wang, Xiao Chen, Xin Jiang, Qun Liu}

\address{
  Huawei Noah's Ark Lab, Shenzhen, China
  }
\email{zhengnianzu@huawei.com, dengliqun.deng@huawei.com}

\begin{document}
\ninept
\maketitle
\begin{abstract}
Mispronunciation detection and diagnosis (MDD) is a popular research focus in computer-aided pronunciation training (CAPT) systems. End-to-end (e2e) approaches are becoming dominant in MDD. 
However an e2e MDD model usually requires entire speech utterances as input context, which leads to significant time latency especially for long paragraphs. 
We propose a streaming e2e MDD model called CoCA-MDD. We utilize conv-transformer structure to encode input speech in a streaming manner. A coupled cross-attention (CoCA) mechanism is proposed to integrate frame-level acoustic features with encoded reference linguistic features. CoCA also enables our model to perform mispronunciation classification with whole utterances. 
The proposed model allows system fusion between the streaming output and mispronunciation classification output for further performance enhancement. 
We evaluate CoCA-MDD on publicly available corpora. CoCA-MDD achieves F1 scores of 57.03\% and 60.78\% for streaming and fusion modes respectively on L2-ARCTIC. For phone-level pronunciation scoring,  CoCA-MDD achieves 0.58 Pearson correlation coefficient (PCC) value  on SpeechOcean762.



\end{abstract}
\section{Introduction}
\label{sec:intro}
Mispronunciation detection and diagnosis (MDD) is a key technology in computer-assisted pronunciation training (CAPT) systems \cite{rogerson2021computer}. In recent years, the emergence of end-to-end (e2e) neural models promises potential performance improvement, leading to new research interests for e2e MDD models \cite{zhang2021text,9414451,wu2021transformer,feng2020sed,leung2019cnn} from both research and commercial communities. 

\begin{figure}[t]
	\centering
	\includegraphics[width=\linewidth]{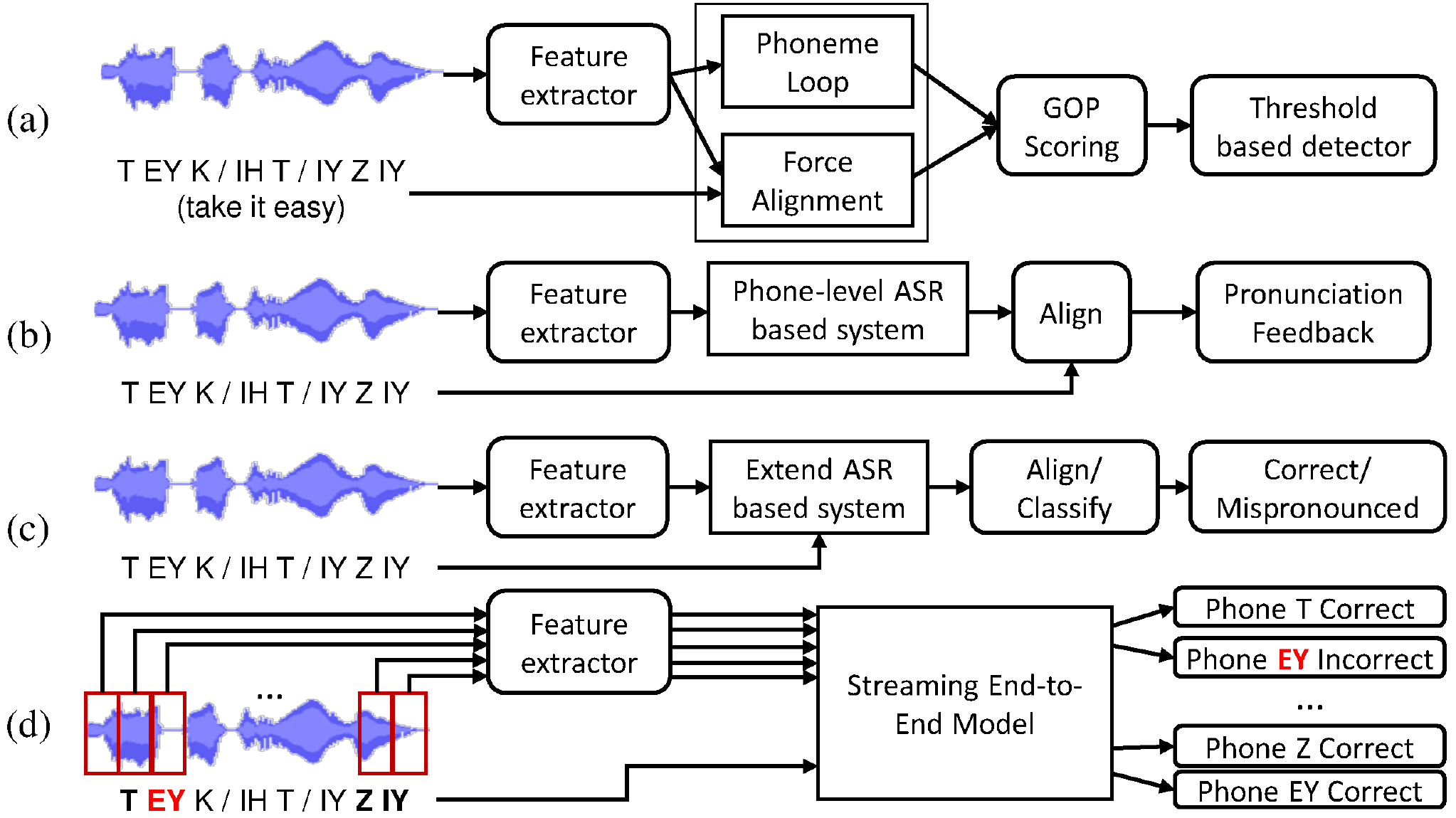}

	\caption{An overview of MDD frameworks. (a) GOP based method, (b) ASR based model, (c) Extended ASR based model, (d) Our streaming framework.}
	\label{fig:illustration}
	\vspace{-5mm}
\end{figure}

Different MDD frameworks are illustrated in Fig. \ref{fig:illustration}. Traditional goodness-of-pronunciation (GOP) \cite{witt2000phone, Huang2017ATL,Dong2019NormalizationOG} based methods (a) usually require a long pipeline of multiple steps. These steps such as forced-alignment and phonological rule matching are prone to error accumulation, leading to inferior MDD performance. 
Automatic Speech Recognition (ASR) based approaches (b) \cite{Hu2015ImprovedMD,Maqsood2017ACS} utilize acoustic encoder to convert input speech into recognized phones. Mispronunciation is detected by aligning recognized phones with phonetic transcription converted from reference text.  The authors in \cite{leung2019cnn} apply a convolution neural network and recurrent neural network (CNN-RNN) as acoustic encoder and Connectionist Temporal Classification (CTC) criterion as training objective. Performance of ASR based methods improves significantly compared with the traditional methods. 
Recent studies suggest that injecting text information during acoustic modeling (c) improves MDD performance \cite{feng2020sed, zhang2021text, fu2021full, 9414451}. SED-MDD \cite{feng2020sed} extends the work in \cite{leung2019cnn} by applying text encoder to encode linguistic features. The encoded linguistic features are merged with acoustic features through attention mechanism for further context modeling. The authors in \cite{zhang2021text} apply a transformer network to implement the similar text-encoding strategy. Their results demonstrate that utilization of prior text prompt improves detection performance. Self-supervised learning (SSL) methods which learn context representation from unlabeled data are explored
in \cite{peng2021study,Yang201920Pronunciation}. SSL allows the same or better detection performance with fewer labelled training data. 

Current MDD systems require reasonable amount of computation. These systems usually start to return MDD feedback only after a user finishes recording of the entire utterance. Without additional engineering efforts, such as operating in blocks, there is considerable amount of delay in returning MDD feedback when the recording is in paragraph-length.
For an ideal CAPT system \cite{Menzel2001InteractivePT}, learners would expect a system to response immediately once they start speaking. The immediate feedback helps the learners to improve sustained attention for the rest of paragraph. A modification for this purpose is to apply streaming architecture to acoustic encoder in e2e models (b) or (c). However,  streaming models usually lead to degraded ASR results and hence lower MDD accuracy due to lack of future context. 

\begin{figure*}[htb]
	\centering
	\begin{subfigure}[b]{0.66\textwidth}
	\includegraphics[scale=0.62]{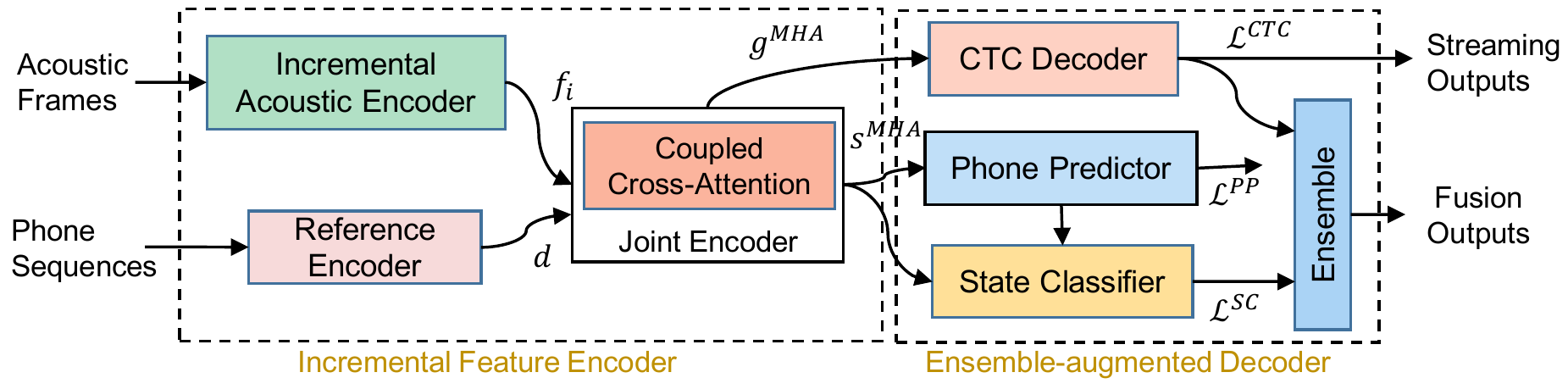}
	\caption{CoCA-MDD}
	\end{subfigure}
	\hfill
	\begin{subfigure}[b]{0.31\textwidth}
	\includegraphics[scale=0.38]{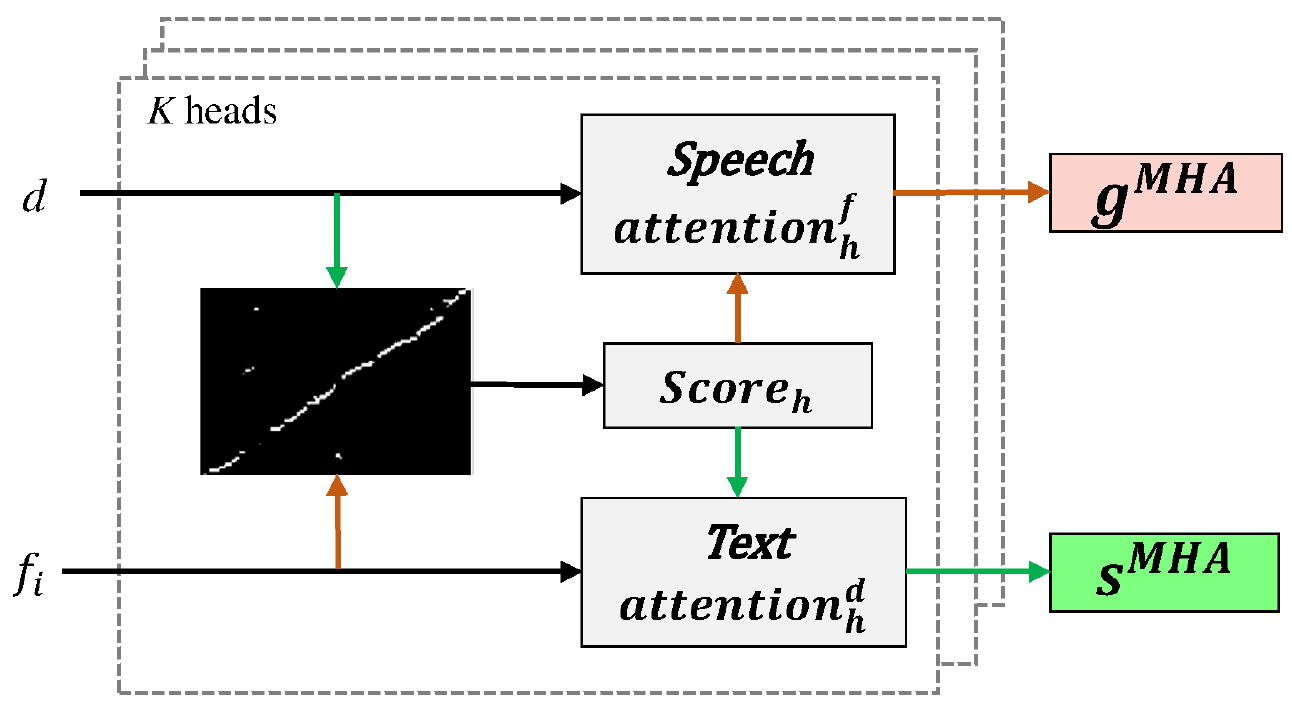}
    \caption{Coupled cross-attention}
	\end{subfigure}
	\caption{(a) Illustration of the proposed CoCA-MDD, which consists of an incremental feature encoder and an ensemble-augmented decoder. (b) Structure of Coupled cross-attention. The upper part is speech attention, which is input to CTC decoder. The text attention is shown in lower part and is used in phone predictor and state classifier.}	
	\label{fig:archi}
\end{figure*}

We propose a streaming e2e MDD framework named CoCA-MDD as illustrated in Fig. \ref{fig:illustration}(d) to alleviate the problem. Motivated by the success of streaming ASR \cite{He2019StreamingES, Huang2020ConvTransformerTL, Li2020TowardsFA}, we apply conv-transformer blocks \cite{Huang2020ConvTransformerTL} as streaming acoustic encoder to encode input speech. To inject given text prompt to the e2e model, we propose a coupled cross-attention (CoCA) mechanism. CoCA allows flexible alignment between input text prompt and streaming acoustic features. 
We further close the performance gap between streaming and offline MDD by applying system fusion when entire speech utterance is available.

Phone-level pronunciation scoring is another useful metric for pronunciation assessment of language learners. We show that CoCA-MDD is convertible to pronunciation scoring task given the manually annotated scores.

This paper is organized as follows. We first describe our proposed CoCA-MDD in the next section. We present the experimental setup and evaluation results in Section 3. Finally we conclude this work in Section 4. 

\section{Proposed Method}
The detail of CoCA-MDD  is shown in Fig. \ref{fig:archi}. CoCA-MDD is composed of an incremental feature encoder and an ensemble-augmented decoder, which are described as follows. 

\subsection{Incremental Feature Encoder (IFE)}
The incremental feature encoder (IFE) contains three parts, i.e., 
an incremental acoustic encoder (IAE), a reference encoder (RE), and a joint encoder (JE). We use the conv-transformer based encoder as described in \cite{Huang2020ConvTransformerTL} to implement our IAE module. Instead of low frame rate of 80~ms with 3 conv-transformer blocks \cite{Huang2020ConvTransformerTL}, IAE operates at 40~ms frame rate with 2 blocks for better speech granularity. Benefited from interleaved transformers and convolution layers, IAE processes acoustic features frame-by-frame with a small look-ahead window (60~ms for CoCA-MDD) for modeling future context. The RE module takes reference phone strings converted from given text prompt by an open-source grapheme-to-phoneme (G2P) \cite{g2pe} as input. We construct the RE module with a network of two bidirectional transformer layers. As the text prompt is known beforehand, the output of RE module is pre-computed before audio recording.
JE consists of a coupled cross-attention (CoCA) layer. JE takes 
linguistic features output by RE and acoustic features from IAE as inputs. 
As shown in Fig. \ref{fig:archi}(b), CoCA consists of 2 multi-head cross-attentions with shared score map. CoCA is formulated as,
\begin{align}
    score_h &= \text{softmax}(\frac{f_i W_{h}^{Q}(d W_{h}^{K})^T}{\sqrt{m}}) \label{eqn:score} \\
    attention_h^{f} &= score_h \cdot (d W_{h}^{V}) \label{eqn:attnh} \\
    g^{MHA} &= \text{concat}_{h\in H}(attention_h^f)W_f^O \label{eqn:attnh2} \\
    attention_h^{d} &= \text{softmax}(score_h^{T}) \cdot (f W_{h}^{QV}) \label{eqn:attns} \\
    s^{MHA} &= \text{concat}_{h\in H}(attention_h^d)W_d^O \label{eqn:attns2}
\end{align}

\noindent where $h$ is denoted as the index of attention heads with a total number of $H$ heads. $m$ is the dimension of attention head. $W_{h}^{K}$, $W_{h}^{V}$ are projection matrices of RE outputs $d$. $W_{h}^{Q}$ is projection matrix of incremental speech feature $f_i$. 
After finishing speech recording, $W_{h}^{QV}$ is projection matrix of accumulated acoustic features $f$. $W_{f}^{O}$ and $W_{d}^{O}$ are output weights of speech attention and text attention respectively.
Concatenation operator is denoted as $\text{concat}(\cdot)$.

The speech attention $attention^f_h$ shown in upper part of Fig.\ref{fig:archi}(b) takes incremental speech feature $f_i$ as query for 
related linguistic features from $d$ as indicated by red arrows. The text attention $attention^d_h$ computes linguistic features by integrating related acoustic features. The two attentions are conditioned on the same matching score $score_h$, which is computed online with currently processed acoustic frames. The speech attention output $g^{MHA}$ is computed frame-by-frame as the whole linguistic context $d$ is known as prior information. The text attention output $s^{MHA}$  is computed with the entire utterance.
The on-the-fly generation of $g^{MHA}$ allows streaming phone recognition. The full-context output $s^{MHA}$ performs final sentence-based mispronunciation detection. 

\subsection{Ensemble-augmented Decoder (EAD)}
A CTC decoder, a phone predictor (PP) and a state classifier (SC) are included in ensemble-augmented decoder (EAD). 
The CTC decoder consists of a unidirectional transformer and one-layer fully connected feed-forward network of 512 hidden units to predict phones from $g^{MHA}$. Pronunciation mistakes are obtained by aligning recognized phones with given phone reference using Needleman–Wunsch algorithm \cite{needleman1970general}. 

Both PP and SC take $s^{MHA}$ as input. PP predicts phone labels. SC classifies whether the phones are pronounced correctly. Each of PP and SC consists of a stack of two bidirectional transformer layers followed by a multi-layer perception (MLP). 
PP uses a softmax layer for final phone prediction. SC takes a logistic layer for output.
SC further takes the output of second transformer of PP as intermediate representation. We observe that adding the intermediate representation to $s^{MHA}$ improves the performance of SC. Since $s^{MHA}$ requires full speech context, both PP and SC do not support streaming output and require entire speech utterances as input. 

\subsection{Training Objective}
The training objective of CoCA-MDD consists of three aspects: CTC-loss $\mathcal{L}_{CTC}$ \cite{Alex2006CONNECTION} for the CTC decoder, cross-entropy loss $\mathcal{L}_{PP}$ for PP and binary cross-entropy loss $\mathcal{L}_{SC}$ for SC. The three decoder are jointly trained in multi-task learning. 
The training objective $\mathcal{L}_{CTC}$ is negative log-likelihood on speech-to-text data $\{x_i, y_i\}$ from data $S$ as the loss function.
\begin{equation}
    \mathcal{L}_{CTC} = - \mathbb{E}_{\langle x,y \rangle \in S} \log P(y|x)
\end{equation}
\noindent where $x$ and $y$ is speech sequence and ground-truth pronounced  phonetic transcription. The $\mathcal{L}_{PP}$ and $\mathcal{L}_{SC}$ are defined as,
\begin{align}
    \mathcal{L}_{PP} &= -\frac{1}{M} \sum_{t=1}^{M}\alpha_t p_{t}\log q_{t}\label{eqn:ce} \\
    \mathcal{L}_{SC} &= -\frac{1}{M}\sum_{t=1}^{M} \alpha_t \bigg[e_t\log s_t + (1-e_t)\log (1-s_t)\bigg] \label{eqn:bce}
\end{align}
\noindent where $M$ is the length of the reference phones and $\alpha_t$ is defined as,
\begin{equation}
    \alpha_t = \begin{cases} 
    \alpha & \text{if $e_t$ = 1} \\
    1 & \text{otherwise} \\
    \end{cases}
\end{equation}
\noindent We introduce $\alpha$ to alleviate class imbalance problem in which there are fewer mispronounced pairs in training data. The targets $p_t$ of PP and $e_t$ of SC are obtained by aligning ground-truth pronounced phones $y$ of speech with the given phone sequence input to RE, where $e_t$ = 1 means that the reference phone is labelled as mispronunciation. The predicted phones from PP are denoted as $q_t$. The probability predicted by SC is denoted as $s_t$. These losses are combined for the objective of model training as following,
\begin{equation}
    \mathcal{L} = \mathcal{L}_{CTC} + \beta \mathcal{L}_{SC} + \gamma \mathcal{L}_{PP} \label{eqn:totalloss}
\end{equation}
where hyper-parameters $\alpha$, $\beta$, and $\gamma$ are set to be 5, 1 and 0.5 based on simple grid search in our experiments.
\subsection{MDD System Fusion}
We further employ a system fusion scheme to combine the predictions for final MDD decision. The ensemble is based on the following rule. For each phone in the reference, when the corresponding SC output is classified as mispronunciation, we view the phone as mispronounced even if the result from the CTC decoder agrees with the reference. The final MDD decision is overridden with $serr$. Otherwise, the final MDD decision always follows the output of the CTC decoder. We apply this rule to improve recall rate 
from streaming MDD decision.  An example is given in Table \ref{tab:ensem}. 
In this example, SC probability $s_t>0.5$ is considered as mispronounced.
Reference phones are phones converted from the text prompt by G2P. Pronounced phones are phones actually pronounced in the recorded speech. 
In this example, 
the word ``she" is not presented in the reference. SC does not output any probability.  The word is pronounced in the speech and is recognized by the CTC decoder. Thus the corresponding phones appear in the fusion result. 
The SC probability on the phone ``EH" of the word ``bed" is greater than 0.5. The phone is considered as mispronounced. The fusion MDD decision becomes \textit{serr}. 
SC sometimes make mistakes. The phone ``T" in the word ``went" is mis-classified as mispronounced, leading to false rejection.
The last phone ``D" in the word ``bed" is missed by the CTC decoder. SC probability is smaller than 0.5. The final MDD decision follows the CTC decoder according to the rule. The missed error is not recovered.


\begin{table}[!t]
\centering
\caption{
The ensemble scheme of CTC decoder and SC module for CoCA-MDD inference
}
\setlength{\tabcolsep}{3pt}
\begin{adjustbox}{width=1\linewidth}
\begin{tabular}{r|cc|cccc|cc|ccc}
\hline
Sentence           & \multicolumn{2}{c|}{\textit{(she)}} & \multicolumn{4}{c|}{went} & \multicolumn{2}{c|}{to} & \multicolumn{3}{c}{bed} \\
Reference Phones                  & -             & -          & W           & EH   & N    & T         & T         & UW          & B        & EH    & D          \\
Pronounced Phones              & SH          & IY         & W           & EH   & N    & T         & T         & UW          & B        & EY     & D          \\ \hline
CTC decoder (streaming)       & SH          & IY         & W           & EH   & N     & T        & T         & UW          & B        & EH    & -         \\ 
SC  probability $s_t$              & -             & -          & 0.0          & 0.0   & 0.0  & 0.63     & 0.0      & 0.4           & 0.0      & 0.92   & 0.44       \\ 
CoCA-MDD (fusion)              & SH          & IY         & W           & EH   &  N   & serr        & T        & UW          & B        & serr     & -          \\ \hline
\end{tabular}
\end{adjustbox}
\label{tab:ensem}
\end{table}


\begin{figure}[th]
	\centering
	\includegraphics[width=0.48\textwidth]{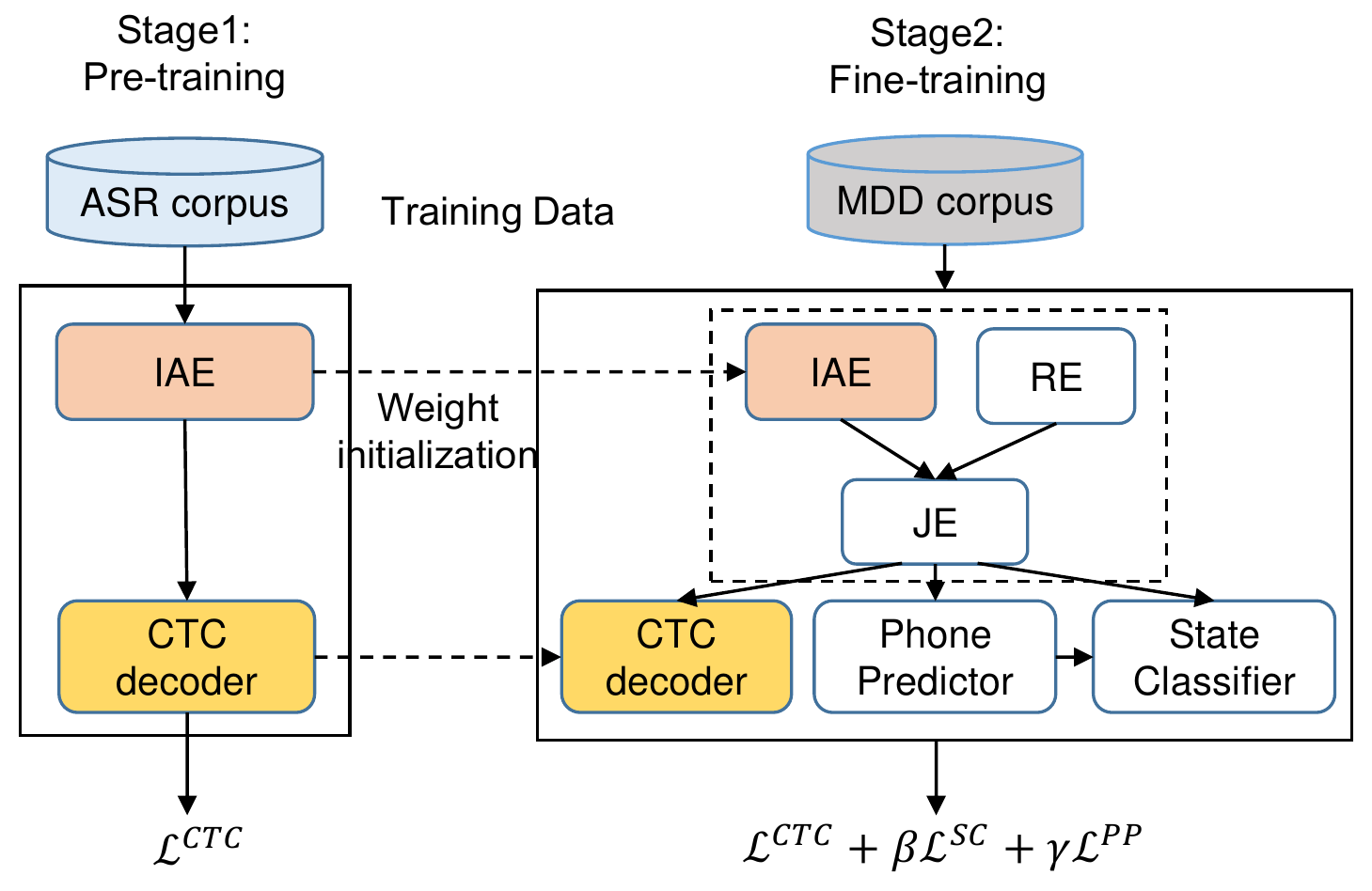}
	\caption{The two-stage training scheme in CoCA-MDD.}
	\label{fig:training}
\end{figure}

\section{Experiments}
\subsection{Experimental Setup}
\noindent \textbf{Dataset} 
We conduct our experiments 
on 3 publicly available corpora, TIMIT \cite{garofolo1993darpa}, L2-ARCTIC  \cite{zhao2018l2}, and SpeechOcean762 \cite{zhang2021speechocean762}.
TIMIT contains recordings of 630 speakers of 8 US English dialects. All the recordings are well-labeled with 61-phone transcription.
L2-ARCTIC is a non-native English corpus. 
We perform our experiments with version 5 of the corpus. 
The corpus consists of recordings from 24 speakers (12 males and 12 females) of six different first languages (Hindi, Korean, Mandarin, Spanish, Arabic and Vietnamese).
Phone-level transcription is available with a 48-phone set, plus an additional \textit{err} symbol for unclear pronunciation. 
We follow the data processing pipeline 
of \cite{fu2021full,peng2021study}. 
We first convert the transcriptions of the two corpora into 39-phone set according to CMUDict \cite{cmudict}. Then we take the recordings of six speakers (NJS, TLV, TNI, TXHC, YKMK, ZHAA) from L2-ARCTIC as test set. The remaining 18 speakers of L2-ARCTIC and TIMIT are used as training set.  There are 7.2~h speech data for training and 0.88~h L2-ARCTIC data for testing in the MDD task.
We further evaluate phone-level pronunciation scoring with SpeechOcean762 \cite{zhang2021speechocean762}. The corpus consists of 5000 English utterances from 250 non-native speakers, with two sets of 2500 utterances for training and testing respectively. The corpus provides manually annotated phone-level accuracy by 5 experts in range of $[0,1,2]$, where 0 is incorrect or missed and 2 is correct. We normalize the scores into $[0,0.5,1]$.

\begin{table*}[!t]
\centering
\caption{MDD results of proposed CoCA-MDD. True rejection is further analysed with correct diagnosis and error diagnosis.  
}
\setlength{\tabcolsep}{3pt}
\begin{adjustbox}{width=1\textwidth}
\begin{tabular}{c|c|c|c|c|c|c|c}
\hline
\multirow{3}{*}{Models} & \multicolumn{2}{c|}{Correct pronunciation}                                 & \multicolumn{3}{c|}{Mispronunciation}                                 & \multirow{3}{*}{F1(\%)} & \multirow{3}{*}{PER(\%)} \\ \cline{2-6}
                        & \multirow{2}{*}{True Acceptance} & \multirow{2}{*}{False Rejection} & \multirow{2}{*}{False Acceptance} & \multicolumn{2}{c|}{True Rejection}   &                         &                          \\ \cline{5-6}
                        &                              &                                  &                               & Correct Diag.          & Error Diag.  &                         &                          \\ \hline
w2v2.0-XLSR+TIMIT \cite{peng2021study}       & 94.30\%(24273)               & 5.70\%(1467)                     & 41.80\%(1783)                 & 70.72\%(1756)          & 29.28\%(727) & 60.44\%                 & 16.20\%                  \\ \hline
CoCA-MDD (streaming)                     & 95.34\%(24517)      & 4.66\%(1197)            & 48.99\%(2102)                 & 80.95\%(1772) & 19.05\%(417) & 57.03\%                 & 11.84\%         \\ \hline
\end{tabular}
\end{adjustbox}
\label{result:mdd}
\end{table*}

\noindent \textbf{Attention configuration} 
All multi-head attentions in transformer layers and CoCA layer consist of 6 heads for each attention. The dimension of each head is 64. The sizes of input and feed-forward layers are 384 and 1536 respectively. 

\noindent \textbf{Data augmentation} 
There are fewer mispronounced examples in training data. The CTC decoder, PP and SC are prone to overfitting with examples of correct pronunciation, hence biasing towards the reference phone input from RE. We augment the phone reference by randomly inserting, deleting or substituting phones to simulate mispronunciation.

\noindent \textbf{Model training} 
We consider the model architecture of CoCA-MDD as an extension of a CTC-based phone-level acoustic model (AM), with an addition of PP and SC. We first perform supervised training to the mentioned CTC-based AM. As shown in Fig.\ref{fig:training}, both CoCA-MDD and the AM share the same IAE and CTC decoder configurations. Rather than training the entire CoCA-MDD from scratch, we find that initializing IAE and CTC decoder with the weights from the trained AM improves 
training stability. The IAE probably learns discriminative acoustic representations during AM training, reducing phone recognition error in this stage. We refer CTC-based AM training to as pre-training and CoCA-MDD training to as fine-tuning. 
We pre-train the model with TIMIT and L2-ARCTIC
in Stage 1. For the MDD task, we fine-tune with L2-ARCTIC. For phone-level pronunciation scoring, we fine-tune with SpeechOcean762 in Stage 2. Note that for MDD task, we label the SC target $e_t=1$ for mispronunciation. For pronunciation scoring, $e_t$ follows the normalized manually annotated scores.

\noindent \textbf{Performance Evaluation}
For MDD task, we follow the evaluation metrics adopted in \cite{fu2021full,peng2021study,leung2019cnn,zhang2021text}. For correct pronunciation, true acceptance ($TA$) indicates correct predictions, while false rejection ($FR$) denotes failure to accept correct pronunciation. In mispronunciation, false acceptance ($FA$) indicates the models mis-classified mispronounced phones as correct, while true rejection ($TR$) indicates that the models detect the mispronounced phones successfully. The F-Measure (F1) is calculated as $2\times precision \times recall / (precision + recall)$, where $precision$ is $TR/(FR+TR)$ and $recall$ is $TR/(TR+FA)$ respectively. The performance of phone recognition is another key factor for MDD. We use phone error rate (PER) as the metric. For phone-level pronunciation scoring, the performance is measured with Pearson Correlation Coefficient (PCC) between predicted scores and reference scores. 

\subsection{Evaluation on Phone Recognition}
Accurate phone decoding is helpful to MDD problem.
The PER results are shown in the last column of Table \ref{result:mdd}. CoCA-MDD (streaming) corresponds to the output from the CTC decoder, which achieves PER of $11.84\%$. 
The result demonstrates the effectiveness of the text-acoustic modeling with CoCA. 


\begin{table}[tb]
\centering
\caption{Performance of CoCA-MDD on Recall, Precision and F1 metrics.}
\vspace{-2mm}
\setlength{\tabcolsep}{3pt}
\begin{adjustbox}{width=1\linewidth}
\begin{tabular}{c|ccc}
\hline
Models              & Recall(\%) & Precision(\%) & F1(\%) \\ \hline
GOP \cite{Yan2021EndtoEndMD}                 & 35.42         & 52.88      & 42.42         \\ 
CTC-ATT \cite{yan2020end}             & 46.57         & \textbf{70.28}      & 56.02         \\ 
CNN-RNN-CTC+VC \cite{fu2021full}      & 56.04         & 56.12      & 56.08         \\ 
w2v2.0-XLSR+TIMIT \cite{peng2021study} & 58.20         & 62.86      & 60.44         \\
\hline
CoCA-MDD (streaming)                 & 51.01         & 64.65      & 57.03         \\ 
CoCA-MDD (fusion)        & \textbf{67.49}         & 55.29      & \textbf{60.78} \\ \hline
\end{tabular}
\end{adjustbox}
\label{tab:t2}
\vspace{-3mm}
\end{table}

\subsection{Evaluation on MDD}
We report the MDD results in Table \ref{result:mdd}. We also include the results from a recent non-streaming MDD system fine-tuned from SSL models \cite{peng2021study} as reference. 
The proposed CoCA-MDD (streaming) achieves true
acceptance and false rejection rates of $95.34\%$ and $4.66\%$  respectively. The results indicate that our model is able to well recognize correct pronunciation. For mispronunciation, our model also achieves reasonable diagnosis accuracy of $80.95\%$ for true rejection. The trade-off of CoCA-MDD (streaming) is high false acceptance rate. We suspect that under streaming configuration, the CTC decoder is able to predict correct phones due to inherit phone language model in IAE and biased linguistic cues from RE.
The high false acceptable rate is alleviated
by our fusion scheme. As shown in Table \ref{tab:t2}, CoCA-MDD (fusion) improves the recall rate from $51.01\%$ to $67.49\%$, and achieves F1-score of $60.78\%$. However, there is a trade-off with lower precision rate ($64.65\% \rightarrow 55.29\%$), with more false rejection after system fusion. The trade-off should be adjustable with the threshold of SC probability. 

We further list the performance of four other non-streaming MDD approaches in Table \ref{tab:t2}. The statistics in Table \ref{tab:t2} suggest that CoCA-MDD (streaming) is among the same performance level as other non-streaming approaches. Note that although all the approaches test on L2-ARCTIC, direct comparison may not be appropriate as different systems are tuned to widely different operating points.



\subsection{Phone-level Pronunciation Scoring}
Our baseline model is a neural network based goodness of pronunciation method (GOP-NN) \cite{HU2015154}. GOP-NN requires a deep neural network to extract GOP-based features. A support vector regressor (SVR) is trained using GOP-based features to predict phone-level scores. System details and results of SpeechOcean762 is published in \cite{zhang2021speechocean762}.

CoCA-MDD directly utilizes SC output probability as phone-level pronunciation scores. The results of pronunciation scoring is shown in Table \ref{tab:t3}. CoCA-MDD achieves PCC of 0.58, which is significantly better than the baseline. Note that CoCA-MDD for pronunciation scoring and MDD can be trained end-to-end at the same fine-tuning stage. CoCA-MDD demonstrates the potential of integrating manually annotated pronunciation scores for MDD system training.

\begin{table}[tb]
\centering
\caption{Performance of CoCA-MDD on phone-level pronunciation scoring}
\vspace{-1mm}
\begin{tabular}{cc}
\hline
Models     & PCC  \\ \hline
GOP-NN + SVR \cite{zhang2021speechocean762} & 0.45 \\
CoCA-MDD       & 0.58 \\ \hline
\end{tabular}
\label{tab:t3}
\vspace{-4mm}
\end{table}

\section{Conclusion}
To our best knowledge, our work is the first attempt to apply an e2e streaming neural model for mispronunciation detection and diagnosis problem. Our proposed CoCA-MDD employs conv-transformer network for stream acoustic data processing, a coupled cross-attention to fully integrate the linguistic and acoustic cues, and a system fusion of CTC decoder and state classifier trained in multi-task learning to improve the MDD accuracy. The result meets the state-of-the-art counterparts.
We further apply CoCA-MDD for phone-level pronunciation scoring and it achieves significant performance improvement from the baseline. As a future work, we are interested in detecting non-categorical error or distortion.


\bibliographystyle{IEEEtran}
\bibliography{paper}

\end{document}